\documentclass[11pt,fleqn]{article} 

\usepackage{a4wide}

\usepackage{amsmath}

\usepackage{graphicx}
\usepackage{subfigure}

\usepackage[numbers]{natbib}
\newcommand{\footaffil}{NICTA}
\newcommand{\foottitle}{}
\usepackage{fancyhdr,lastpage}
\if@twoside
 \fancyheadoffset[LE,RO]{0cm}
 \fancyfootoffset[LE,RO]{0cm}
\fi
\fancypagestyle{plain}
{
 \fancyhf{}
    
 \fancyhead[L]{}
 \fancyhead[C]{}
 \fancyhead[R]{}
 
 \fancyfoot[L]{}
 \fancyfoot[C]{}
 \fancyfoot[R]{}
     
}

\fancypagestyle{default}
{

 \fancyhf{}
        
 \fancyhead[L]{\textsf{\leftmark}}
 \fancyhead[C]{}
 \fancyhead[R]{\textsf{\rightmark}}
        
 \fancyfoot[L]{\footnotesize\textsf{\footaffil}}
 \fancyfoot[C]{\footnotesize\textsf{\thepage/\pageref{LastPage}}}
 \fancyfoot[R]{\footnotesize\textsf{\foottitle}}
    
}

\fancypagestyle{special}
{

 \fancyhf{}
        
 \fancyhead[L]{\textsf{\leftmark}}
 \fancyhead[C]{}
 \fancyhead[R]{}

 \fancyfoot[L]{\footnotesize\textsf{\footaffil}}
 \fancyfoot[C]{\footnotesize\textsf{\thepage/\pageref{LastPage}}}
 \fancyfoot[R]{\footnotesize\textsf{\foottitle}}
    
}

\fancypagestyle{frontmatter}
{

 \fancyhf{}
        
 \fancyhead[L]{}
 \fancyhead[C]{}
 \fancyhead[R]{}

 \fancyfoot[L]{}
 \fancyfoot[C]{\footnotesize\textsf{\thepage}}
 \fancyfoot[R]{}
    
}

\newcommand{\GS}{\ensuremath{\mathcal{G}}}
\newcommand{\G}{\ensuremath{\mathcal{G}^d}} 
\newcommand{\sNS}{\mathcal{N}} 
\newcommand{\sES}{\mathcal{E}}
\newcommand{\sTS}{\mathcal{T}} 
\newcommand{\sSS}{\mathcal{S}} 
\newcommand{\sAS}{\mathcal{A}}
\newcommand{\sN}{\mathcal{N}^d} 
\newcommand{\sE}{\mathcal{E}^d} 
\newcommand{\sT}{\mathcal{T}^d}
\newcommand{\sS}{\mathcal{S}^d} 
\newcommand{\sA}{\mathcal{A}^d} 
\newcommand{\horiz}{\mathcal{H}}

\newcommand{\flow}[2][]{\varphi_{{#2}}^{{#1}}}
\newcommand{\binf}[1]{x_{{#1}}}
\newcommand{\bmcf}[2]{x_{{#1}}^{{#2}}}

\newcommand{\nin}[1]{\delta^-(#1)}
\newcommand{\nout}[1]{\delta^+(#1)}
\newcommand{\nins}[1]{\delta_0^-(#1)}
\newcommand{\nouts}[1]{\delta_0^+(#1)}
\newcommand{\UB}{\textsc{ub}}
\newcommand{\sto}{\mbox{s.t.}\quad }

\newcommand{\et}[1]{\ensuremath{s_{#1}}}
\newcommand{\eu}[1]{\ensuremath{u_{#1}}}
\newcommand{\ef}[1]{\ensuremath{\bar{f}_{#1}}}

\newcommand{\nd}[1]{\ensuremath{d_{#1}}}
\newcommand{\nf}[1]{\ensuremath{\bar{f}_{#1}}}

\newcommand{\xp}[1]{x_{#1}}

\newcommand{\sP}[1][]{\Omega_{#1}}

\newcommand{\ctrref}[2]{\eqref{lp:#1}-\eqref{lp:#2}}

\usepackage{array,tabularx,booktabs,multirow}

\newcolumntype{R}{>{\raggedleft\arraybackslash}X}%

\newcommand{\head}[1]{\multicolumn{1}{c}{\textbf{#1}}}

\usepackage{algorithm}
\usepackage{algorithmicx}
\usepackage{algpseudocode}

\algblockx[ParForall]{ParForall}{EndParForall}%
[2]{\textbf{parallel\ forall}\ #1\ \textbf{in}\ #2\ \algorithmicdo}%
{\textbf{end forall}}
\algblockx[ParFor]{ParFor}{EndParFor}%
[2]{\textbf{parallel\ for}\ #1\ \textbf{to}\ #2\ \algorithmicdo}%
{\textbf{end for}}

\newcommand{\lref}[1]{(line~\ref{#1})}

\newcommand{\fn}[2]{\ensuremath{\mathtt{#1}\left(#2\right)}}

\makeatletter
\newcommand*{\overt}[1]{$\overline{\hbox{#1}}\m@th$}
\makeatother

\usepackage[bookmarks,pdftex,pdftitle={A Conflict-Based Path-Generation Heuristic for Evacuation Planning},
              pdfauthor={Victor Pillac, Pascal Van Hentenryck, Caroline Even},
              pdfkeywords={Evacuation planning and scheduling, regional evacuation, disaster management, conflict-based path generation}]{hyperref}

\usepackage{authblk}

\title{A Conflict-Based Path-Generation Heuristic for Evacuation Planning}

\date{September 2013}

\author[1]{Victor Pillac\footnote{Corresponding author: victor.pillac@nicta.com.au}}
\author[1,2]{Pascal Van Hentenryck}
\author[1,3]{Caroline Even}

\affil[1]{NICTA, Victoria Research Laboratory\authorcr
 Melbourne, Australia}

\affil[2]{Department of Computing and Information Systems, University of Melbourne\authorcr
 Melbourne, Australia}

\affil[3]{Ecole des Mines de Nantes\authorcr
 Nantes, France}
\renewcommand{\foottitle}{NICTA 2013}
\renewcommand{\footaffil}{Pillac, Van Hentenryck, Even}

\begin{document}

\pagestyle{plain}
\maketitle


\begin{center}
\noindent\rule{\textwidth}{.5pt}
\end{center}

\begin{abstract}
  Evacuation planning and scheduling is a critical aspect of disaster management
and national security applications. This paper proposes a conflict-based path-generation 
approach for evacuation planning. Its key idea is
to generate evacuation routes lazily for evacuated areas and to optimize the
evacuation over these routes in a master problem. Each new path is generated to
remedy conflicts in the evacuation and adds new columns and a new row in the
master problem. The algorithm is applied to massive flood scenarios in the
Hawkesbury-Nepean river (West Sydney, Australia) which require evacuating in the
order of 70,000 persons. The proposed approach reduces the number of variables
from 4,500,000 in a Mixed Integer Programming (MIP) formulation to 30,000 in the
case study. With this approach, realistic evacuations scenarios can be solved
near-optimally in real time, supporting both evacuation planning in strategic,
tactical, and operational environments.  

\paragraph{Keywords:}
Evacuation planning and scheduling, regional evacuation, disaster management, conflict-based path generation
\end{abstract}

\begin{center}
\noindent\rule{\textwidth}{.5pt}
\end{center}

\newpage
\pagestyle{default}
\section{Introduction}
Natural disasters, such as hurricanes, floods, and bushfires, affect
numerous populated areas and may endanger the lives and welfare
of entire populations. Evacuation orders are some of the most important
decisions performed by emergency services: They ensure the safety of
persons at risk by instructing them to evacuate the threatened region,
be it a building (e.g., fire), a neighborhood (e.g., industrial
hazard), or a whole region (e.g., flood). Evacuation planning also
arises at strategic, tactical, and operational levels.  At a
strategic level, the goal is to design evacuation plans for specific
areas and possible threats (e.g., evacuation plans for the
surroundings of a nuclear power plant). At a tactical level, the goal
is to design evacuation plans for an area facing an incoming threat
(e.g., evacuation of a flood plain following high precipitations).
Finally, at the operational level, the goal is to schedule an
evacuation, possibly adjusting the evacuation plan in real-time as the
threat unfolds.

Existing work in evacuation planning typically consider free-flow models in
which evacuees are dynamically routed in the network. However, free-flow models
do not conform to existing evacuation methodologies in which evacuated nodes
are assigned specific evacuation routes (see, for instance, \cite{HNFESP2005}).

In contrast, this paper presents an evacuation algorithm that follows
recommended evacuation methodologies: It generates evacuation routes
for evacuated nodes and uses a lexicographic objective function that
first maximizes the number of evacuees and then postpones the
evacuation as much as possible. The algorithm can be used for
strategic and tactical planning and is fast enough to operate in
real-time conditions, even for large evacuations.

From a technical standpoint, the algorithm can be broadly
characterized as a Conflict-Based Path-Generation Heuristic (CPG for
short), which shares some characteristics with column generation approaches.
As in column generation, we decompose the problem by considering separately the
generation of evacuation paths (subproblem) and their selection (master problem). 
However, a challenge of our application is the spatio-temporal nature of the problem: 
one evacuation path corresponds to multiple paths in the spatio-temporal graph modeling the actual scheduling of the evacuation. 
Therefore, a single path introduces multiple
columns in the master problem, which increases the complexity the pricing of
a new path. To tackle this limitation, the path-generation subproblem aims at finding
a path of least cost under constraints, where the edge costs are
derived from the conflicts and congestion in the incumbent
evacuation.

The CPG algorithm was evaluated on real-scale, massive flood scenarios
in the Hawkesbury-Nepean river (West Sydney, Australia) which require
evacuating in the order of 70,000 persons. Experimental results
indicate that the CPG algorithm generates high-quality solutions in
real time. On small instances, where optimal solutions can be found,
the CPG algorithm finds optimal or near-optimal solutions. On real-scale
instances, the results show that the CPG algorithm is capable of
evacuating the entire Hawkesbury-Nepean region even if the population grows
by 40\%. The solution quality of the CPG algorithm can be bounded by
free-flow models that provide (optimistic) upper bound on solution
quality: The results indicate that its solutions (in terms of
evacuees) are within 13\% of free-flow models even if the population
is increased by 200\%. Finally, microscopic traffic simulations show that the
solution produced by CPG are robust to different evacuees behaviours.

The remainder of this paper is organized as follows: Section
\ref{section:formulation} formulates the evacuation planning problem, Section
\ref{sec:relw} reviews related work, Section \ref{sec:mf} presents three
solution approaches, while Section \ref{sec:sched} discusses two formulations to
postpone the evacuation as much as possible. Section \ref{sec:comp} compares the
performance of the proposed approaches on a set of realistic instances. Finally,
Section \ref{sec:conc} concludes this paper.

\section{Problem Formulation}
\label{section:formulation}

\begin{figure}[htb]
  \centering
  \subfigure[Evacuation Scenario\label{fig:evac-scen}]{\includegraphics[width=.48\textwidth,page=1]{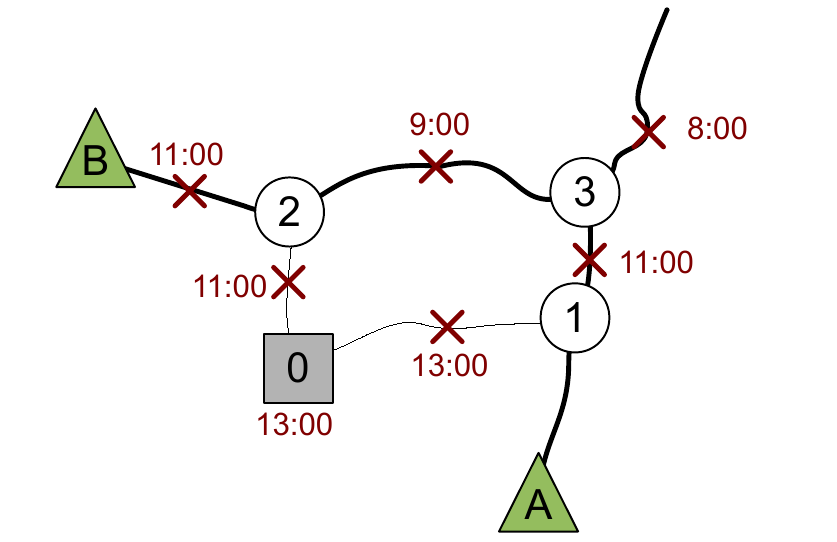}}
  \hfill
  \subfigure[Evacuation Graph\label{fig:evac-graph}]{\includegraphics[width=.48\textwidth,page=2]{fig1_evacuation_network}}
\vspace{-0.7mm}
  \caption{Modeling of an Evacuation Planning Problem.}
\label{fig:evac}
\vspace{-0.4mm}
\end{figure}

Figure \ref{fig:evac} illustrates a general evacuation
scenario. Fig. \ref{fig:evac-scen} presents an evacuation scenario
with one evacuated node (0) and two safe nodes (A and B). In this
example, the evacuated node 0 has to be evacuated by 13:00,
considering that certain links become unavailable at different times
(for instance, $(2,3)$ is cut at 9:00). This evacuation scenario can
be represented as a graph $\GS=(\sNS=\sES\cup\sTS\cup\sSS,\sAS)$ where
$\sES$, $\sTS$, and $\sSS$ are the set of evacuated, transit, and safe
nodes respectively, and $\sAS$ is the set of edges. Each evacuated
node $i$ is characterized by a number of evacuees $\nd{i}$ and an
evacuation deadline $\nf{i}$ (e.g., 20 and 13:00 for node 0
respectively), while each edge $e$ is associated with a triple
$(\et{e},\eu{e},\ef{e})$, where $\et{e}$ is the travel time, $\eu{e}$
is the capacity, and $\ef{e}$ is the time at which the edge becomes
unavailable.

\begin{figure*}[htb]
  \centering
  \includegraphics[width=.75\textwidth]{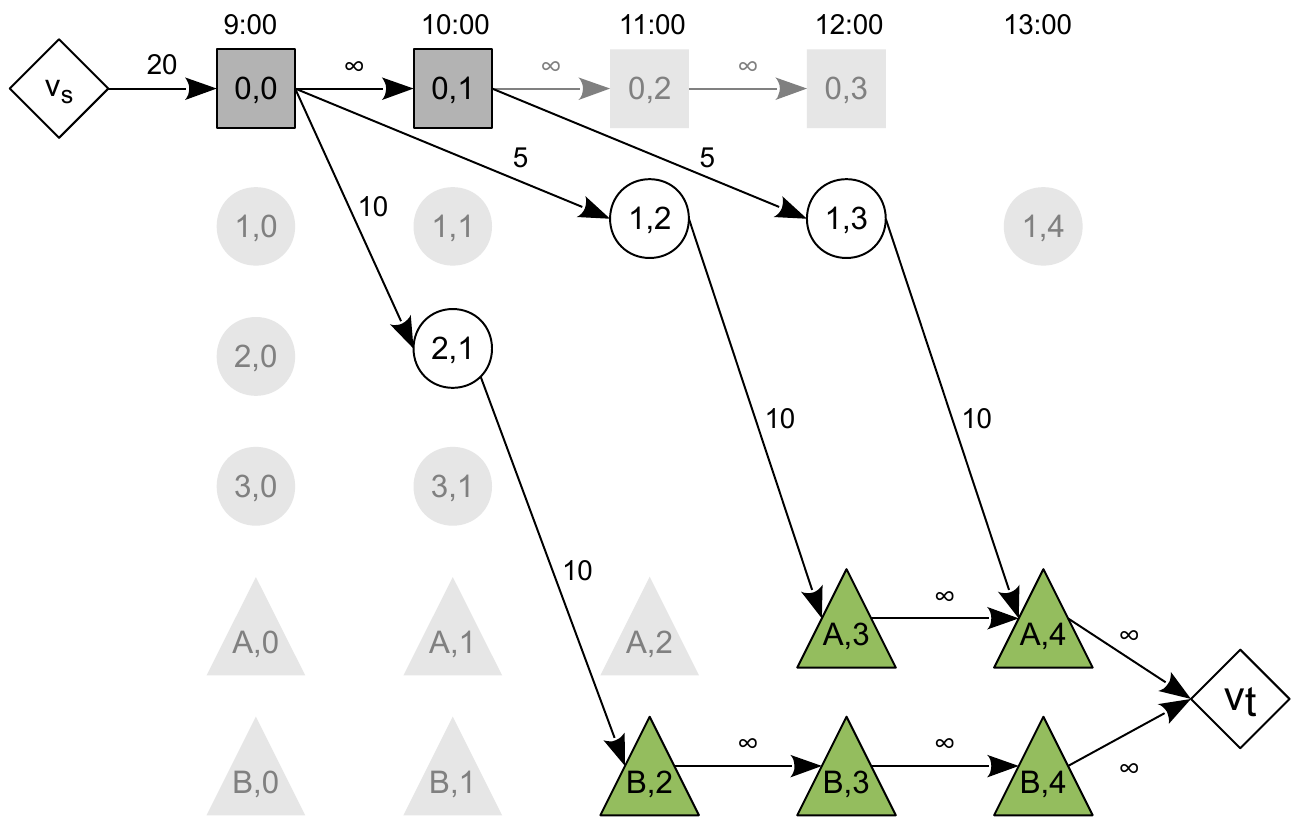}
  \caption{Time-Expanded Graph for the Evacuation Scenario With 1-hour Time Steps.}
\label{fig:evac-dyn}
\vspace{-0.4mm}
\end{figure*}

A common way to deal with the space-time aspects of evacuation
problems is to discretize the planning horizon into time steps of
identical length, and to work on a \emph{time-expanded graph}, as
illustrated in Fig. \ref{fig:evac-dyn}. This graph
$\G=(\sN=\sE\cup\sT\cup\sS,\sA)$ is constructed by duplicating each
node from $\sNS$ for each time step. For each edge $(i,j)\in\sAS$ and
for each time step $t$ in which edge $(i,j)$ is available, an edge
$(i_t,j_{t+\et{(i,j)}})$ is created modeling the transfer of evacuees
from node $i$ at time $t$ to node $j$ at time $t+\et{(i,j)}$. In
addition, edges with infinite capacity are added to model the evacuees
waiting at evacuated and safe nodes. Finally, all evacuated nodes (resp.
safe nodes) are connected to a virtual super-source $v_s$ 
(super-sink $v_t$), modeling the inflow (outflow) of
evacuees. Note that some nodes may not be connected to either the
super-source or super-sink (in light gray in this example), and can
therefore be removed to reduce the graph size. The problem is then to
find a flow from $v_s$ to $v_t$ that models the movements of evacuees
in space and time.

In this study, we consider a single threat scenario from which we derive the 
time when each evacuated node must be evacuated, and the time at which
edges are closed. In addition, we ignore the dynamics of the actual evacuees 
movements, and assume that the edge capacity is fixed and does not depend on the flow. This is a necessary simplification that is compensated by the fact that edge
capacities are set to ensure non-congested flow. In addition, we follow the
practice in the field of emergency services operations and assume that all
evacuees from a same node will be evacuated to the same safe node 
following a single path.

The objective is to first ensure that all evacuees reach a safe node, and then
to delay the evacuation as much as possible. This second objective is motivated
by the type of threat considered: we assume that evacuees safety is only threatened
after the evacuation deadline. Therefore, it is of practical interest to evacuate
them as late as possible, as this leaves more time to potentially refine 
the threat scenario and hence avoids unnecessary evacuations.
The decisions that need to be made for each evacuated are the following: which safe
node to evacuate to, which path to follow to reach the selected safe node, and 
how to schedule the departures over the horizon. Finally, the global evacuation
plan and schedule must respect the capacity of the network, and ensure that 
no evacuee travels on an edge that has been cut. 

\section{Related work}\label{sec:relw}

According to \citet{Hamacher2001}, evacuation planning can be tackled
using either \emph{microscopic} or \emph{macroscopic}
approaches. Microscopic approaches focus on modeling and simulating
the evacuees individual behaviors, movements, and
interactions. Macroscopic approaches, such as the three presented in
this study, aggregate evacuees and model their movements as a flow in
the evacuation graph.


To the best of our knowledge, all models
are free-flow.  A significant number of contributions attempt to
solve flow problems directly derived from the time-expanded
graph. For instance, \citet{Lu2003,Lu2005} propose three heuristics to design an
evacuation plan with multiple evacuation routes per evacuated node,
minimizing the time of the last evacuation.  The
authors show that in the best case the proposed heuristic is able to solve randomly
generated instances of up to 50,000 nodes and 150,000 edges in under 6
minutes. \citet{Liu2007a} propose a Heuristic Algorithm for Staged
Traffic Evacuation (HASTE), a similar algorithm that generates
evacuation routes and schedule the evacuation of evacuated nodes in
sequence. The main difference between HASTE and the previous
algorithms is that it relies on a Cell Transmission Model (CTM)\citep{Daganzo1994}
to model more accurately the flow of evacuees.

Acknowledging that all evacuated nodes may not be under the same level
of threat, \citet{Lim2012} consider a short-notice regional evacuation
maximizing the number of evacuees reaching safety weighted by the severity
of the threat.  The authors propose two solution approaches to solve
the problem, and present computational experiments on instances
derived from the Houston-Galveston region (USA) with up to 66 nodes,
187 edges, and a horizon of 192 time steps.

Other authors have focused on modeling more accurately the
transportation network.  For example, \citet{Bretschneider2011,Bretschneider2012} 
present a free-flow
mathematical model that describes in detail the street network and, in
particular, the lane configuration at intersections of the network.
They present computational experiments on generated instances with a
grid topology of up to 240 nodes, 330 edges, and considering 150 times
steps.  \citet{Bish2013} present a model based on a CTM that assigns a single
evacuation path to each evacuated node. Computational results include
instances with up to 13 evacuated nodes, 2 safe nodes, and 72 edges.

Finally, dynamic aspects of evacuation have also been considered. For
instance, \citet{Lin2008} present a time expanded graph in which they
allow for time-dependent attributes such as varying capacity or
demand. The authors apply their findings on a case study considering the
evacuation of a 11-floor building with approximately 60 nodes, 100
edges, and 60 time steps.

Microscopic approaches include the work by \citet{Richter2013} 
who challenge two assumptions generally made: The existence of a
central planning entity with global knowledge, and the ability of this
entity to communicate order to evacuees. They propose a decentralized
decision making approach supported by smartphones and mobile
applications. We note however that our target applications, such as
evacuations for floods and hurricanes, use central decision making and
have the time and ability to communicate their decisions.


Column generation is an optimization technique which consists in
considering only a subset of columns in a master problem and then
iteratively generating columns of negative reduced cost (assuming
minimization) by solving a pricing subproblem. It has been widely used
to solve large-scale MIP problems, and we refer the interested reader
to the book by \citet{Desaulniers2005} and the study
by \citet{Luebbecke2005} for a recent
review of techniques and applications of column generation. In
particular, it has been used to solve multi-commodity network flow
problems (MCNF) \citep{Alvelos2000}, integer MCNF \citep{Barnhart2000},
origin-destination MCNF \citep{Barnhart1997}, and MCNF with side
constraints on paths \citep{Holmberg2003}. However, a distinctive
feature of evacuation planning is the dependency between paths in the
time-expanded network. More precisely, a commodity (i.e., evacuees
from a specific evacuated node) can only follow paths that correspond
to the same physical path (sequence of edges in the evacuation
graph). Therefore classical MCNF approaches cannot be applied
directly, as one path in the evacuation model introduces multiple
variables in the master problem. In addition, it is worth noting that
heuristic column generation have mainly focused on solving the pricing
subproblem heuristically. In contrast, our approach does not consider
the pricing problem explicitly, but heuristically generates new
paths. Similar ideas were also used by \citet{Coffrin2011}, and, to 
a lesser extent, by \citet{Massen2012}.

\section{Proposed Approaches}
\label{sec:mf}

In this section we present three approaches of increasing practical
relevance to the decision maker. The first approach is based on
free-flow models and is used as a bound to evaluate the other two
models. The second approach uses a Mixed Integer Programming (MIP)
formulation to ensure that evacuees from a same evacuated node follow a single
evacuation path. Finally, the third approach is our conflict-based heuristic
path generation algorithm.

\subsection{Free-Flow Model}
\label{ssec:ff}
The Free-Flow model (FF) assumes that evacuees can flow freely in the
graph. From a practical perspective, this corresponds to an ideal case
in which the evacuees are given the order to evacuate and are then
dynamically routed in the graph. In other words, evacuees do not know
in advance the path to follow in the graph. Although of limited
practical importance, this model provides a bound for more advanced
evacuation planning.

\begin{figure*}[htb]
\begin{align}
\max\quad & \Phi = \sum_{e\in\nin{v_t}} \flow{e} \label{lp:mf-obj}\\
\sto & \sum_{e\in\nin{i}}\flow{e} - \sum_{e\in\nout{i}}\flow{e} = 0 & \forall i
\in\sN\setminus\{v_s,v_t\} \label{lp:mf-flow}\\
& \flow{e} \leq \eu{e} 		& \forall e\in\sA \label{lp:mf-ub}\\
& \flow{e} \geq 0 	& \forall e\in\sA \label{lp:mf-end}
\end{align}
\caption{The Free-Flow model (FF).}\label{lp:ff}
\end{figure*}

Figure \ref{lp:ff} presents the Free Flow model (FF). $\flow{e}$ is the flow of evacuees on edge $e\in\sA$, and $\nin{i}$ ($\nout{i}$) is the set of inbound (outbound) edges adjacent to node $i$. The objective \eqref{lp:mf-obj} is to maximize the number of evacuees reaching a safe node. Constraints \eqref{lp:mf-flow} ensure the conservation of the flow in the graph, while constraints \eqref{lp:mf-ub} enforce the capacity on edges. The demand of the evacuations nodes is modeled implicitly as a capacity on the edges connecting them to the super-source ($u_{(v_s,i)}=d_i$).
Note that the flow of evacuees is considered as a continuous quantity. This is motivated by the fact that the considered number of evacuees and edge capacity are already approximations of the reality, thus a unitary granularity is not necessary.

\subsection{Restricted-Flow Model}
\label{ssec:od}
The restricted-Flow (RF) model enforces the constraint that each node is
evacuated along a single path. They can be thought of as a form of
multi-commodity flows but the spatio-temporal nature of evacuation
planning introduces some key differences discussed later in the paper.
The formulation of the model is interesting in that it expresses some
of its constraints using the {\em evacuation graph} and others using
the {\em time-expanded graph}.  In general, the evacuation graph is
used to enforce constraints on paths, while the time-expanded graph is
instrumental in stating the flow constraints. Obviously, both graphs are
necessary for some constraints. For simplicity, we denote $e_0$ the edge in the evacuation
graph that corresponds to edge $e$ in the time-expanded graph,
projecting out the time information.

Figure \ref{lp:rf} presents the RF model.
It introduces a binary variable $\bmcf{e_0}{k}$ which is
equal to 1 if and only if edge $e_0\in\sAS$ belongs to the evacuation
path for evacuated node $k$, and a continuous variable $\flow[k]{e}$
equal to the flow of evacuees from evacuated node $k$ on edge
$e\in\sA$. 
The objective \eqref{lp:od-obj} is to maximize the number of evacuees
reaching a safe node. Constraints \eqref{lp:od-st-onepath} ensure that
exactly one path is used to route the flow coming from a same
evacuated node in the evacuation graph, while constraints
\eqref{lp:od-st-flow} ensure the continuity of the path. Constraints
\eqref{lp:od-flow} ensure the flow conservation through the
time-expanded graph. Constraints \eqref{lp:od-ub} enforce the capacity
of each edge in the time-expanded graph. Constraints \eqref{lp:od-ubp}
ensure that there is no flow of evacuees coming from an evacuated node
$k$ if edge $e$ is not part of the evacuation path for $k$.

\begin{figure*}[htb]
\begin{align}
\max\quad & \Phi = \sum_{e\in\nin{v_t}}\sum_{k\in\sES} \flow[k]{e} \label{lp:od-obj}\\
\sto & \sum_{e_0\in\nouts{k}}\bmcf{e_0}{k} = 1 & \forall k \in \sES \label{lp:od-st-onepath} \\
& \sum_{e_0\in\nins{i}}\bmcf{e_0}{k} - \sum_{e_0\in\nouts{i}}\bmcf{e_0}{k} = 0 & \forall k \in \sES, i \in \sTS \label{lp:od-st-flow} \\ 
& \sum_{e\in\nin{i}}\flow[k]{e} - \sum_{e\in\nout{i}}\flow[k]{e} = 0 & \forall i \in\sN\setminus\{v_s,v_t\},  k \in \sES \label{lp:od-flow}\\
& \sum_{k\in\sES}\flow[k]{e} \leq u_{e} 		& \forall e\in\sA \label{lp:od-ub}\\
& \flow[k]{e} \leq u_{e}*\bmcf{e_0}{k} 		& \forall e\in\sA,  k \in \sES \label{lp:od-ubp}\\
& \flow[k]{e} \geq 0 	& \forall e\in\sA,  k \in \sES\\
& \bmcf{e}{k} \in \{0,1\} & \forall e\in\sA,  k \in \sES \label{lp:od-end}
\end{align}
\caption{The Restricted-Flow model (RF).}
\label{lp:rf}
\end{figure*}

\subsection{Conflict-Based Heuristic Path Generation}
The main drawback of the RF model is its complexity, both in terms of
number of variables and constraints. As the experimental results
demonstrate, the model is computationally intractable even for small instances.
To address this issue, we propose a conflict-based heuristic path generation approach (CPG) that
separates the generation of evacuation paths from the scheduling of
the evacuation.

Algorithm \ref{algo:cg} gives an overview of the approach. First, an
initial set of paths $\sP'$ is generated \lref{algo:cg-init} and a
master problem is solved to find an evacuation schedule that maximizes
the number of evacuees reaching safety \lref{algo:cg-mpi}. The
procedure then identifies \emph{critical} evacuated nodes
\lref{algo:cg-crit}, which are not fully evacuated, or evacuated early. 
This information is later used to generate new paths
\lref{algo:cg-gen}. Finally the scheduling problem is solved including
the newly generated paths \lref{algo:cg-mp}. The last four steps are
repeated for a given number of iterations or until a predefined number
of non-improving iterations has been reached \lref{algo:cg-stop}.

\begin{algorithm*}[htb]
\newcommand{\sol}{\mathcal{S}}
 \begin{algorithmic}[1]
   \Require $\GS$ the evacuation graph, $\G$ the time-expanded graph.
   \Ensure $\sol$ the best solution found
   \State $\sP'\gets \fn{generatePaths}{\GS,\emptyset,\sES,\emptyset}$ \label{algo:cg-init}\Comment{Subproblem}
   \State $\sol\gets\fn{scheduleEvacuation}{\Omega',\GS,\G}$\label{algo:cg-mpi}\Comment{Master problem}
   \While{stopping criterion not met}\label{algo:cg-stop}
     \State $\sES_c\gets\fn{findCriticalEvacuatedNodes}{\sol}$\label{algo:cg-crit}
     \State $\sP'\gets\sP'\cup\fn{generatePaths}{\GS,\sP',\sES_c,\sol}$\label{algo:cg-gen} \Comment{Subproblem}
     \State $\sol\gets\fn{scheduleEvacuation}{\sP',\GS,\G}$\label{algo:cg-mp}\Comment{Master problem}
   \EndWhile
   \State \Return $\sol$
 \end{algorithmic}
\caption{The Conflict-Based Path Generation.}
\label{algo:cg}
\vspace{-0.4mm}
\end{algorithm*}


The master problem can be solved using a mixed integer program. Let
$\sP$ be the set of all feasible paths between evacuated nodes and
safe nodes and $\sP[k]$ be the subset of evacuation paths for
evacuated node $k$.  We define a binary variable $\xp{p}$ which takes
the value of 1 if and only if path $p\in\sP$ is selected, a continuous
variable $\flow[t]{p}$ representing the number of evacuees to start
evacuating on path $p$ at time $t$, and a continuous variable
$\flow{k}$ accounting for the number of non-evacuated evacuees in node
$k$. In addition, we denote by $\omega(e)$ the subset of paths that
contain edge $e$ and by $\tau_p^e$ the number of time steps required to
reach edge $e$ when following path $p$. Finally, we note $\horiz_p
\subseteq \horiz$ the subset of time steps in which path $p$ is
usable, and $u_p$ the capacity of path $p$.

\begin{figure*}[htb]
\begin{align}
\max\quad & \sum_{p\in\sP}\sum_{t\in\horiz_p}\flow[t]{p}  \label{lp:rcg-obj}\\
\sto& \sum_{p\in\sP[k]}\xp{p} = 1		  & \forall k\in\sES \label{lp:rcg-sp} \\
& \sum_{p\in\sP[k]}\sum_{t\in\horiz_p} \flow[t]{p} + \flow{k} = \nd{k}  & \forall k\in\sES \label{lp:rcg-dem}\\
& \sum_{\substack{p\in\omega(e) \\ t-\tau_p^e\in\horiz_p}}\flow[t-\tau_p^e]{p} \leq \eu{e}       & \forall e\in\sAS, t\in\horiz\label{lp:rcg-ub}\\
& \sum_{t\in\horiz_p}\flow[t]{p} \leq |\horiz_p|\xp{p} u_p & \forall p\in\sP\label{lp:rcg-pflow}\\
& \flow[t]{p} \geq 0             & \forall p\in\sP, t \in \horiz_p\\
& \flow{k} \geq 0                & \forall k \in\sES\\
& \xp{p}\in\{0,1\}               & \forall p\in\sP
\end{align}
\caption{The evacuation scheduling problem (CPG-MP).}\label{lp:cpgmp}
\end{figure*}

Figure \ref{lp:cpgmp} presents the evacuation scheduling problem CPG-MP.
The objective \eqref{lp:rcg-obj} maximizes the total flow of evacuees,
which is equivalent to the number of evacuees reaching safety.
Constraints \eqref{lp:rcg-sp} ensure that exactly one path is selected
for each evacuated node, while constraints \eqref{lp:rcg-dem} account
for the number of evacuated and non-evacuated evacuees.  Constraints
\eqref{lp:rcg-ub} enforce the capacity on the edges of the graph.
Finally, constraints \eqref{lp:rcg-pflow} ensures that there is no
flow on paths that are not selected. It is interesting to observe that the
master model does not use a variable for each edge $e$ and time step $t$.
Instead, it reasons in terms of variables $\flow[t]{p}$ which indicate
how many evacuees leave along path $p$ at time $t$.

\begin{figure*}[t]
 \centering
 \includegraphics{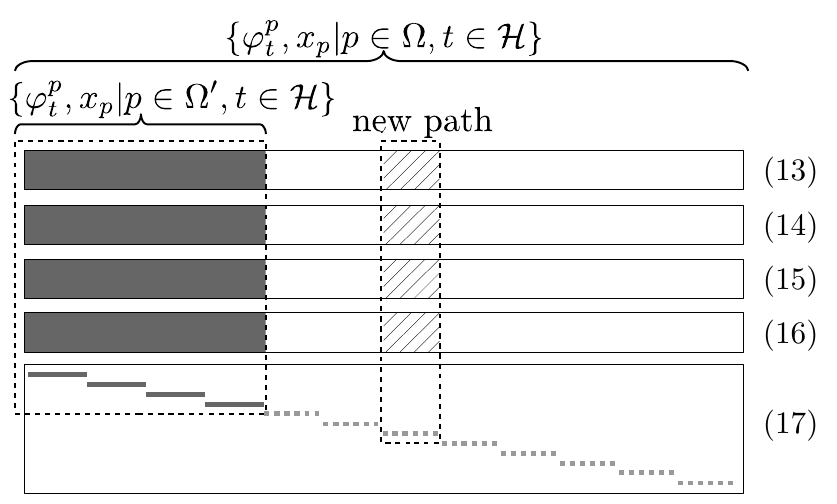}
 \caption{The Structure of the Evacuation Scheduling Master Problem Matrix.}
\label{fig:cg-mp}
\vspace{-0.4mm}
\end{figure*}

In practice, we only consider a subset of evacuation paths
$\sP'\subseteq\sP$ each time we solve CPG-MP.  Fig. \ref{fig:cg-mp}
depicts the structure of the master problem matrix. 
Horizontal blocks represent groups of constraints numbered as in 
Figure \ref{lp:cpgmp}, while the shaded areas represent 
non-null coefficients in the matrix.
Note that each constraint in group \eqref{lp:rcg-pflow} only involves
variables associated with the corresponding path and must be dynamically added 
to the model whenever a new path is considered. 
Nonetheless, a solution of CPG-MP considering the subset of paths
$\sP''\subset\sP'$ is also a feasible when considering the set $\sP'$.
Hence the solution from the previous iteration is used as
starting solution for the current iteration.

Traditionally, the generation of new columns searches for a column of
positive reduced costs (assuming maximization). Because of the
spatio-temporal nature of this application, and the fact that a path
corresponds to multiple columns and introduces a new constraint, we
follow a different approach. We use a conflict-based path
generation which relies on problem-specific knowledge to generate new
columns that will potentially improve the objective function of the
master problem. First, we identify the subset $\sES'\subseteq\sES$ of
critical evacuated nodes, i.e., nodes that are not fully evacuated in the
current solution. Then, we include in $\sES'$ all the evacuated nodes
whose evacuation paths share at least one edge with a node from
$\sES'$.  Finally, we generate new paths for the critical evacuated
nodes $\sES'$ by solving the following multiple-origins,
multiple-destinations shortest path problem: 
\begin{align}
\min\quad & \sum_{k\in\sES'}\sum_{e\in\sAS}c_{e}y_{e}^{k} & \label{lp:pg-obj} \\
\sto & \sum_{e\in\nin{i}}y_e^{k} - \sum_{e\in\nout{i}}y_e^{k} = 0 & \forall i \in \sTS, k \in \sES' \label{lp:pg-flow}\\
& \sum_{e\in\nout{k}}y_{e}^{k} = 1 & \forall k \in \sES' \label{lp:path-unicity}\\
& y_{e}^{k} \in\{0,1\} & \forall k\in\sES', e\in\sAS\label{lp:pg-end}
\end{align}
where $y_e^k$ is a binary variable taking the value of 1 if and only
if edge $e$ belongs to the path generated for evacuated node $k$, and
$c_e$ is the cost assigned to edge $e$. In order to generate diverse
evacuation paths, the cost $c_e$ of edge $e$ is adjusted at each iteration using 
the following linear combination of the edge's travel time $\et{e}$, the number 
of occurrences of $e$ in the current set of paths, and the utilization 
of $e$ in the current solution:
\begin{equation}
c_{e} = \alpha_t\frac{\et{e}}{\max_{e\in\sES}\et{e}} +
 \alpha_c\frac{\sum_{\substack{p\in\sP'\\ e \in p}}1}{|\sP'|} +
 \alpha_u\frac{\sum_{\substack{p\in\sP'\\ e \in p}} \sum_{t\in\horiz_p}\flow[t]{p}}{\eu{e}}\label{eq:c_e}
\end{equation}
where $\alpha_t$, $\alpha_c$, and $\alpha_u$ are positive weights which sum is equal to $1$.

\section{Postponing the Evacuation}
\label{sec:sched}
The three models presented in the previous section share the common
objective of maximizing the number of evacuees reaching safety. However,
in the practical applications considered, stakeholders are also
interested in delaying the evacuation as much as possible. This is
motivated by the fact that, as the disaster unfolds, more information
is available on the nature, extent, and timing of the threat, for
instance with more accurate weather forecasts. Consequently, in this
section we propose two formulations to maximize the time of the first
evacuation, and design evacuation plans that not only guarantee that all
evacuees can reach safety, but are also of practical relevance.

The first approach consists in solving a model maximizing the number
of evacuees reaching safety and then using a post-optimization to
maximize the time of the first evacuation. We illustrate this approach
with the FF model, leading to the FF-E formulation. Similar
transformations are applied to the RF and CPG models, defining the RF-E
and CCG-E formulations. Note that, in the case of CPG-E, new columns can
be generated either when maximizing the number of evacuees or when
maximizing the time of the first evacuation. Let $\binf{t}$ be a
binary variable that takes the value of $1$ if at least
one evacuee leaves any evacuated node at time step $t$. In addition,
we define the continuous variable $\Delta$ as the time of the
first evacuation. We define $\Delta_{\UB}$ as the time at which the
first evacuated node is flooded. The FF-E model is defined as
follows: 

\begin{align}
\max\quad & \Delta  \label{lp:maxs-obj}\\
\sto& \ctrref{mf-flow}{mf-end} & \nonumber \\
& \sum_{e\in\nout{v_s}} \flow{e} \geq \Phi_{max} & \label{lp:maxs-flow}\\
& \sum_{i\in\sE_t}\sum_{e\in\nout{i}}\flow{e} \leq  \binf{t}\sum_{i\in\sE_t} \sum_{e\in\nout{i}}\eu{e} 	 & \forall t\in\horiz \label{lp:maxs-bflow}\\
& \Delta \leq t\binf{t} + (1-\binf{t})\Delta_{\UB} & \forall t\in\{1,\dots,\Delta_{\UB}\} \label{lp:ms-maxs}\\
& \Delta \geq 0 		& \\
& \binf{t} \in \{0,1\} & \forall t\in\horiz
\end{align}

In this model, the objective \eqref{lp:maxs-obj} is to maximize the time of the
first evacuation $\Delta$. Constraint \eqref{lp:maxs-flow} ensures that the flow
is at least the max flow $\Phi_{max}$ found in the first step. Constraints
\eqref{lp:maxs-bflow} define the binary variables $\binf{t}$, constraints
\eqref{lp:ms-maxs} ensure that $\Delta$ is bounded by the time of the first
evacuation.

The drawback of the explicit formulations is that they introduce a
binary variable per time step and require to perform two
optimizations, leading to increased running times. Therefore, we
introduce an implicit method that maximizes the number of evacuees and
the scheduling of the evacuation in a single step. With the purpose of
modeling ``non-evacuated'' evacuees, we add an edge between the last
time-copy of each evacuated node and the super-sink, with zero travel
time and infinite capacity.  The scheduling of the evacuation is achieved by
associating a penalty with edges leaving an evacuated node inversely
proportional to the time slice in which they belong.
Formally, let $c_{(i,j)}$ the cost of edge $(i,j)\in\sA$, defined as:
\begin{equation*}
c_{(i,j)} = \left\{
\begin{array}{l@{\quad}l}
	c_{ne} & \mbox{if } i\in\sE, j=v_t\\
	\frac{H-t(i)}{H} & \mbox{if } i\in\sE, j\in \sT \\
	0 & \mbox{otherwise} 
\end{array}
\right.
\end{equation*}
where $t(i)$ is the time slice of time-node $i$, and $c_{ne}$ is a high penalty for non-evacuated evacuees. The Implicit Free Flow model (FF-I) minimizes the total cost of the flow for all edges subject to the conservation of the flow as follows:
\begin{align}
\min\quad & \sum_{e\in\sA} c_e\flow{e} \label{lp:mfs-obj} \\
\sto& \ctrref{mf-flow}{mf-end} & \nonumber \\
& \flow{(v_s,i)} \geq \nd{i} 	  & \forall i\in\sES
\end{align}  
Similarly to the explicit formulations, this transformation is
extended to the RF and CPG models, leading to the RF-I and CPG-I
formulations.

\section{Computational Experiments}\label{sec:comp}
\subsection{Case Study}
\begin{figure*}[t]
  \subfigure[Map of the area of interest\label{fig:case-map}]{%
    \includegraphics[width=.48\textwidth]{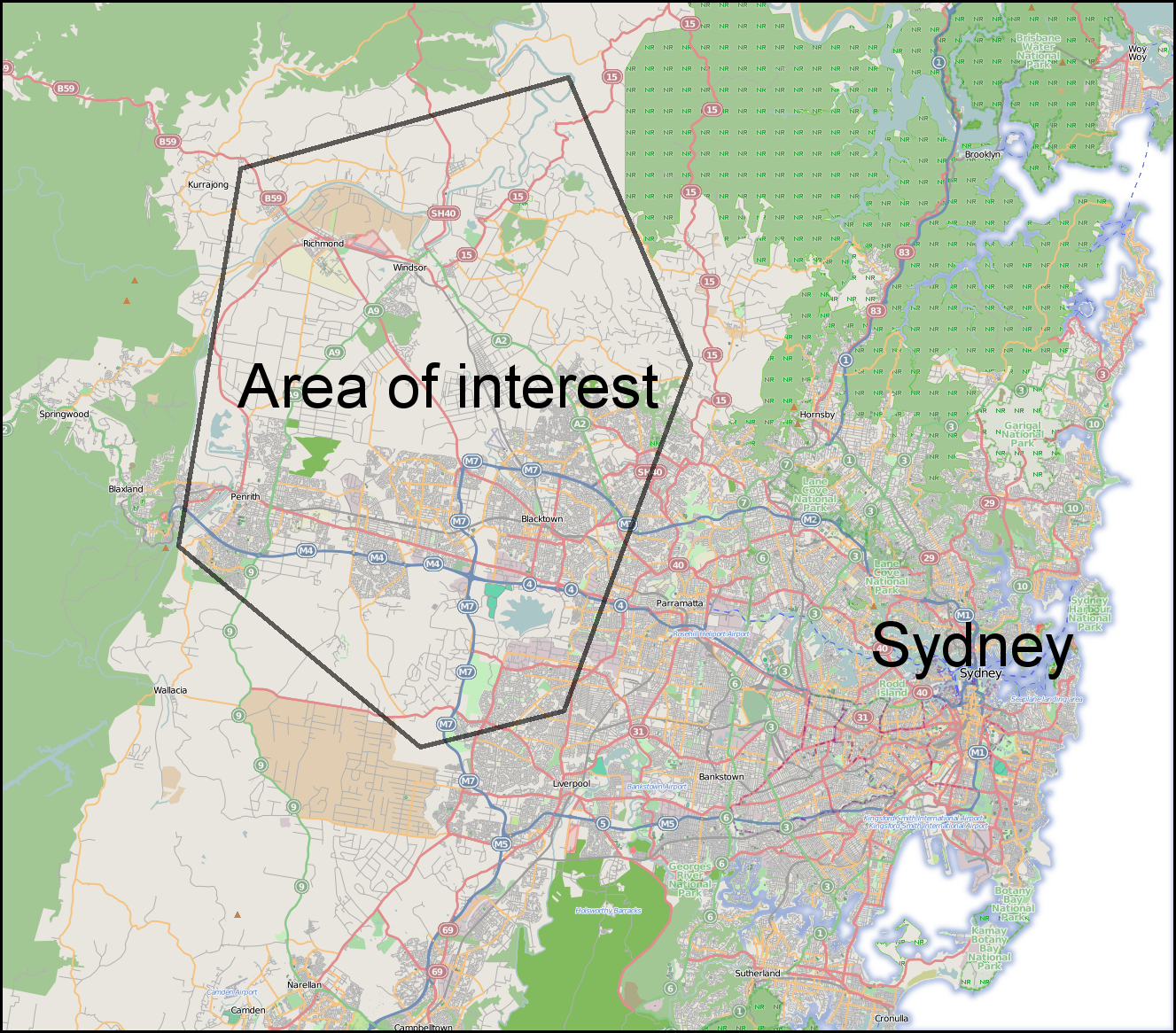}%
  }%
  \subfigure[Evacuation graph for instance HN\label{fig:case-net}]{%
    \includegraphics[width=.48\textwidth]{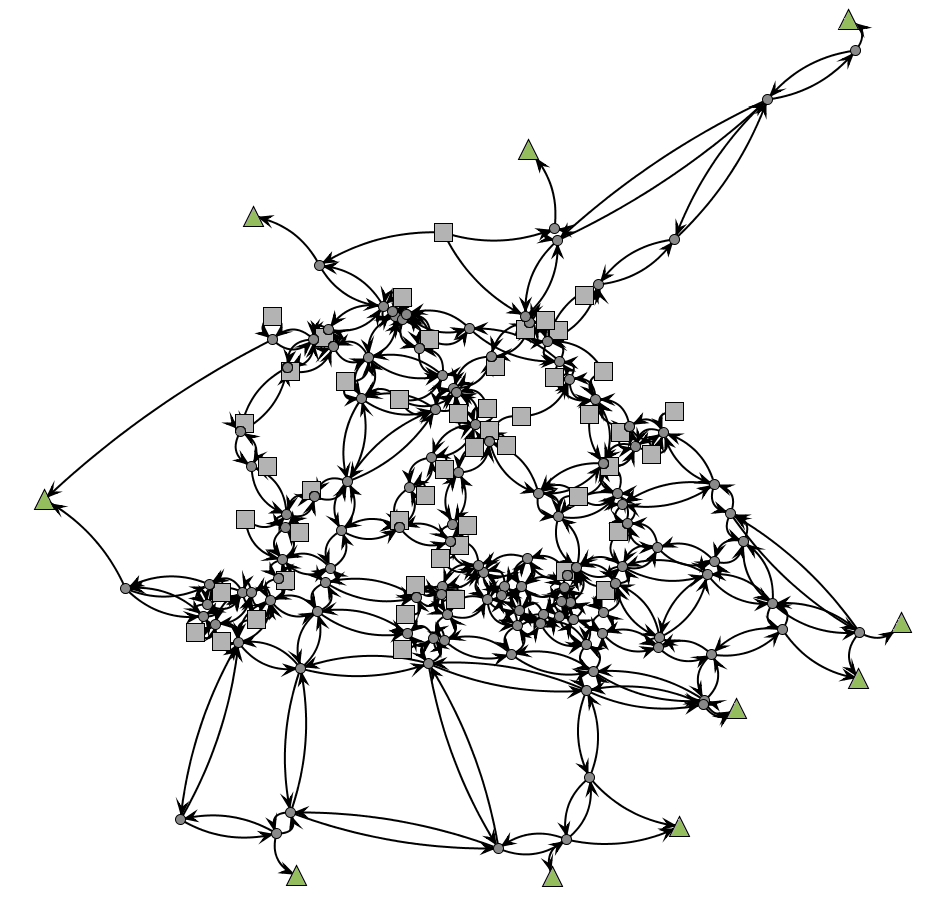}%
  }%
  \caption{Geographical location of the case study}\label{fig:case}
\vspace{-0.4mm}
\end{figure*}

To assess the performance of our algorithms, we considered the
evacuation of the Hawkesbury-Nepean (HN) floodplain, located
North-West of Sydney (see Fig. \ref{fig:case-map}), for which a
1-in-200 years flood will require the
evacuation of 70,000 persons. The resulting evacuation graph,
illustrated in Fig. \ref{fig:case-net}, contains 50 evacuated nodes,
10 safe nodes, 125 transit nodes, and 458 edges. We considered a
horizon of 18 hours with a time step of 5 minutes (starting at 00h00).
The evacuation deadlines and times at which edges are cut were derived
from a flooding scenario similar to the historical 1867 flood \citep{HNFESP2005}.

In addition, we generated two sets of instances based on the case
study. Instances HN-Rn were generated by first selecting the $\mbox{n}\in[2,50]$ 
earliest flooded evacuated nodes, and then reducing the graph by retaining 
only the nodes and edges which are part of the shortest path between each
evacuated node and the closest safe node.
On the other hand, instances HN-Ix have the same evacuation graph as HN with a number of evacuees scaled
by a factor of $\mbox{x}\in[1.1,3.0]$.  All approaches were
implemented using Java 7 and {\sc Gurobi 5.1.1}, and experiments were
conducted on an Ubuntu 12.04 64bits machine with a 2.4Ghz 8-cores
Intel Xeon processor and 32Gb of RAM. Results are an average over 10
runs given the randomized nature of parts of the algorithms and of {\sc Gurobi} internal heuristics.

\subsection[Explicit v.s. implicit formulations]{Comparison of explicit and implicit formulations}
\begin{table*}[t]
\centering
\footnotesize
\begin{tabularx}{\textwidth}{llRRR@{\qquad}RRR}
\toprule
 &  & \multicolumn{3}{c}{\textbf{CPU Time (s)}} & \multicolumn{3}{c}{\textbf{Evacuation Start Time}} \\ 
 \cmidrule(r){3-5} \cmidrule(l){6-8}
\textbf{App.} & \textbf{Instance} & \head{E} & \head{I} & \head{Speedup} & \head{E} & \head{I} & \head{Gap} \\ \midrule
\textbf{FF} & \textbf{HN-R02} & 1.0 & 0.0 & - & 13h35 & 13h35 & +0h00 \\ 
 & \textbf{HN-R03} & 2.0 & 0.0 & - & 14h15 & 14h15 & +0h00 \\ 
 & \textbf{HN-R05} & 0.0 & 0.0 & - & 12h25 & 12h25 & +0h00 \\ 
 & \textbf{HN-R08} & 12.0 & 0.0 & - & 13h10 & 13h10 & +0h00 \\ 
 & \textbf{HN-R10} & 11.0 & 0.0 & - & 13h10 & 13h10 & +0h00 \\ 
 & \textbf{HN-R20} & 39.0 & 0.0 & - & 12h10 & 12h05 & +0h05 \\ 
 & \textbf{HN-R30} & 28.0 & 0.0 & - & 10h05 & 10h00 & +0h05 \\ 
 & \textbf{HN-R40} & 32.0 & 1.0 & 32.0 & 03h15 & 03h10 & +0h05 \\ 
 & \textbf{HN-R50} & 28.0 & 1.0 & 28.0 & 03h15 & 03h10 & +0h05 \\ \midrule
\textbf{RF} & \textbf{HN-R02} & 0.0 & 0.0 & - & 13h35 & 13h35 & +0h00 \\ 
 & \textbf{HN-R03} & 18.0 & 1.0 & 18.0 & 14h00 & 14h00 & +0h00 \\ 
 & \textbf{HN-R05} & 323.0 & 51.0 & 6.3 & 12h25 & 12h05 & +0h20 \\ 
 & \textbf{HN-R08} & \textit{1800.0} & \textit{1800.0} & \textit{-} & \textit{-} & \textit{07h55} & \textit{-}  \\ 
 & \textbf{HN-R10} & \textit{1800.0} & \textit{1800.0} & \textit{-} & \textit{-} & \textit{09h40} & \textit{-}  \\ 
 & \textbf{HN-R20} & \textit{1800.0} & \textit{1800.0} & \textit{-} & \textit{-} & \textit{00h00} & \textit{-}  \\ 
 & \textbf{HN-R30} & \textit{1800.0} & \textit{1800.0} & \textit{-} & \textit{-} & \textit{03h05} & \textit{-}  \\ 
 & \textbf{HN-R40} & \textit{1800.0} & \textit{1800.0} & \textit{-} & \textit{-} & \textit{-} & \textit{-}  \\ 
 & \textbf{HN-R50} & \textit{1800.0} & \textit{1800.0} & \textit{-} & \textit{-} & \textit{-} & \textit{-} \\ \midrule
\textbf{CPG} & \textbf{HN-R02} & 0.1 & 0.0 & - & 13h35 & 13h35 & +0h00 \\ 
 & \textbf{HN-R03} & 0.9 & 0.2 & 4.7 & 13h55 & 13h55 & +0h00 \\ 
 & \textbf{HN-R05} & 2.5 & 1.1 & 2.3 & 12h25 & 12h25 & +0h00 \\ 
 & \textbf{HN-R08} & 7.7 & 1.1 & 6.7 & 12h10 & 11h53 & +0h17 \\ 
 & \textbf{HN-R10} & 11.5 & 1.9 & 5.9 & 12h10 & 11h55 & +0h16 \\ 
 & \textbf{HN-R20} & 13.4 & 10.1 & 1.3 & 11h12 & 11h35 & -0h23 \\ 
 & \textbf{HN-R30} & 14.5 & 8.3 & 1.7 & 09h25 & 09h35 & -0h10 \\ 
 & \textbf{HN-R40} & 9.7 & 2.6 & 3.8 & 03h15 & 03h11 & +0h05 \\ 
 & \textbf{HN-R50} & 10.6 & 2.7 & 3.9 & 03h15 & 03h10 & +0h05 \\ 
\bottomrule
\end{tabularx}
\vspace{2mm}
\caption{Experimental Results on Reduced-Size Instances.}\label{tab:r}
\end{table*}

Table \ref{tab:r} reports the CPU time and evacuation start time for
each of the HN-Rn instances and the Free Flow (FF), Restricted Flow
(RF), and Path Generation (CPG) approaches, using both Implicit (I)
and Explicit (E) formulations. Note that all approaches were able to
evacuate the totality of the evacuees in the reported solutions.
These results highlight the practical limitation of the RF-E and RF-I
approaches, which are not able to terminate in the enforced 30min time
limit for instances with more than 5 evacuated nodes. Although the
RF-I approach is able to find solutions for instances HN-R08 to
HN-R30, this solution is of significantly lower quality than those of
CPG-I.  In addition, it appears that the CPG formulations produce
solutions very similar to those produced by RF for the three smallest
instances and competitive with FF for all instances, which means that
CPG is able to find high-quality solutions with a reduced number of
evacuation paths. Note that the reduced graph of instances HN-Rn 
implies that there exists only a limited number of paths departing each evacuated node, 
hence the similarities between the CPG and FF solutions. Finally, the table
illustrates the benefits of the implicit formulation over the explicit
one, as it significantly reduces the computational times while
maintaining similar solution quality. 
Consequently, in the following, we focus on the implicit formulations.

\subsection{Results on real-size instances}
\begin{table*}[t]
\centering
\footnotesize
\begin{tabularx}{\textwidth}{llRRRRRR}
\toprule
 &  & \head{Num.} &\head{Num.} & \head{Num.} &\head{CPU} & \head{Perc.} & \head{Evac.} \\ 
\textbf{App.} & \textbf{Instance} & \head{Routes} &\head{Cols ($10^3$)} & \head{Rows ($10^3$)} &\head{Time (s)} & \head{Evac.} & \head{Start} \\  \midrule
\textbf{FF-I} & \textbf{HN} & \multicolumn{1}{l}{} & 88 & 32 & 1 & 100\% & 8h35 \\ 
\textbf{} & \textbf{HN-I1.1} &  & 88 & 32 & 1 & 100\% & 8h15 \\ 
 & \textbf{HN-I1.2} &  & 88 & 32 & 2 & 100\% & 7h55 \\ 
 & \textbf{HN-I1.4} &  & 88 & 32 & 2 & 100\% & 7h25 \\ 
 & \textbf{HN-I1.7} &  & 88 & 32 & 2 & 100\% & 7h05 \\ 
 & \textbf{HN-I2.0} &  & 88 & 32 & 3 & 100\% & 6h10 \\ 
 & \textbf{HN-I2.5} &  & 88 & 32 & 3 & 100\% & 5h05 \\ 
 & \textbf{HN-I3.0} &  & 88 & 32 & 3 & 100\% & 3h15 \\  \midrule
\textbf{RF-I} & \textbf{HN} & \multicolumn{1}{l}{} & 4440 & 5655 & 1800 & - & - \\ 
\textbf{} & \textbf{HN-I1.1} &  & 4440 & 5655 & 1800 & - & - \\ 
 & \textbf{HN-I1.2} &  & 4440 & 5655 & 1800 & - & - \\ 
 & \textbf{HN-I1.4} &  & 4440 & 5655 & 1800 & - & - \\ 
 & \textbf{HN-I1.7} &  & 4440 & 5655 & 1800 & - & - \\ 
 & \textbf{HN-I2.0} &  & 4440 & 5655 & 1800 & - & - \\ 
 & \textbf{HN-I2.5} &  & 4440 & 5655 & 1800 & - & - \\ 
 & \textbf{HN-I3.0} &  & 4440 & 5655 & 1800 & - & - \\  \midrule
\textbf{CPG-I} & \textbf{HN} & 140 & 32 & 79 & 15 & 100\% & 3h20 \\ 
 & \textbf{HN-I1.1} & 140 & 32 & 79 & 23 & 100\% & 2h25 \\ 
 & \textbf{HN-I1.2} & 143 & 32 & 79 & 19 & 100\% & 1h50 \\ 
 & \textbf{HN-I1.4} & 143 & 32 & 79 & 25 & 100\% & 0h20 \\ 
 & \textbf{HN-I1.7} & 102 & 23 & 79 & 2 & 98\% & 0h00 \\ 
 & \textbf{HN-I2.0} & 100 & 23 & 79 & 2 & 92\% & 0h00 \\ 
 & \textbf{HN-I2.5} & 110 & 25 & 79 & 3 & 91\% & 0h00 \\ 
 & \textbf{HN-I3.0} & 116 & 26 & 79 & 8 & 87\% & 0h00 \\ 
\bottomrule
\end{tabularx}
\vspace{2mm}
\caption{Experimental Results on Real-Size Instances.}\label{tab:i}
\end{table*}

Table \ref{tab:i} presents computational results for the original HN
instance and the HN-Ix instances for the three approaches and implicit
formulations. The first column reports the number of paths generated,
the second and third give the number of columns and rows in the
MIP, the fourth reports CPU times, the sixth contains the percentage of
evacuees reaching safety, and finally the seventh reports the time of
the first evacuation. As expected, the RF-I approach is unable to find
a feasible solution in the 30min time limit, while the CPG-I can
solve all instances in under 30s. This table also highlights the
dramatic reduction in model size that the CPG-I approach provides, with
a number of columns reduced from 4.4 millions to 32 thousands, and a
number of constraints reduced from 5.7 millions to 79 thousands when
compared with the RF-I approach. Interestingly, the \mbox{CPG-I} model
contains fewer variables that the simpler FF-I model. If we consider
the percentage of evacuees to reach safety, we can note that the FF-I
model is always able to evacuate 100\% of evacuees, while the CPG-I
approach finds optimal solutions for instances up to HN-I1.4 (i.e.,
with a population increased by 40\% with respect too the current
census). The CPG-I approach is still able to evacuate 87\% of the
population in scenarios where the population is increased threefold. 
Finally, CPG-I produces schedules that start much earlier than FF-I. 

These results are particularly compelling considering that the free-flow
models are not realistic and are only useful to provide upper bounds
on solution quality.
Contrary to the reduced instances, there are many paths that the free-flow models can exploit,
making these models significantly over-optimistic.
Although we cannot retrace exactly the paths followed by each evacuee, 
a closer look at instance HN shows that FF-I uses more than one evacuation path
for at least 11 evacuated nodes. In addition, the flow of evacuees is split 
at 18 transit nodes (versus 0 for CPG-I), and FF-I uses 380 edges (82\% of total)
while CPG-I uses only 169 (37\%). 
Finally, applying FF-I to an instance reduced to the nodes and edges present in the 
CPG-I solution yields a evacuation starting at 3h40 (compared to 3h20 for CPG-I),
with 2 evacuated nodes with more than one evacuation path, and a flow split on at least 7 transit nodes.
This illustrates that the gap in performance between the two approaches is due to the fact that FF-I distributes the flow of evacuees from one zone over the entire evacuation graph, which is not realistic in practice.


\subsection{Comparison with upper bound}
Observing that the flow of evacuees is split in a number of transit nodes in FF, we derive a model, namely \overt{FF}, that
prevents this behavior and provides an upper bound for the evacuation time. In this model, we define a binary variable for
each edge of the static graph equal to one if and only if the edge is used. In addition, we add constraints that ensure
that the number of used outbound edges is lower than the number of used inbound edges for each transit node. To prevent
oscillating flows that will open as many edges as required, we add extra constraints that force each edge to be used in a
single direction. Note that both additional constraints are sound from a practical perspective. The same constraints were
also added to the CPG model for comparison, leading to the \overt{CPG} approach.

Table \ref{tab:iub} presents a comparison of \overt{FF} and \overt{CPG} for the HN and HN-Ix instances. Two observations can be made
from the results. First, the new constraints have a dramatic impact on the performance of \overt{FF}, which does not find the
optimal solution in two hours, while they have a limited impact on \overt{CPG} which still terminates under one minute on average.
Second, the schedule of the evacuation produced by \overt{FF} is much closer to \overt{CPG} than FF is to CPG. This supports the claim
that free flow models produce solutions that are overly optimistic when practical constraints are taken into account.
\begin{table*}[h!]
\centering
\footnotesize
\begin{tabularx}{\textwidth}{llRRRRRR}
\toprule
&  & \head{Num.} & \head{Num.} &\head{CPU} & \head{MIP} & \head{Perc.} & \head{Evac.} \\ 
\textbf{App.} & \textbf{Instance}  &\head{Cols ($10^3$)} & \head{Rows ($10^3$)} & \head{Time (s)} & \head{Gap.} & \head{Evac.} & \head{Start} \\  
\midrule
\textbf{\overt{FF}} & \textbf{HN} & 89 & 34 & 7200 & 4.15\% & 100\% & 3h40 \\ 
 & \textbf{HN-I1.1} & 89 & 34 & 7200 & 4.74\% & 100\% & 2h50 \\ 
 & \textbf{HN-I1.2} & 89 & 34 & 7200 & 4.09\% & 100\% & 2h40 \\ 
 & \textbf{HN-I1.4} & 89 & 34 & 7200 & 3.87\% & 100\% & 5h05 \\ 
 & \textbf{HN-I1.7} & 89 & 34 & 7200 & 2.90\% & 100\% & 3h00 \\ 
 & \textbf{HN-I2.0} & 89 & 34 & 7200 & 2.58\% & 100\% & 1h15 \\ 
 & \textbf{HN-I2.5} & 89 & 34 & 7200 & 3.77\% & 100\% & 0h00 \\ 
 & \textbf{HN-I3.0} & 89 & 34 & 7200 & 63.54\% & 96\% & 0h00 \\ 
 \midrule
\textbf{\overt{CPG}} & \textbf{HN} & 45 & 79 & 188 & - & 100\% & 3h15 \\ 
 & \textbf{HN-I1.1} & 20 & 79 & 4 & - & 100\% & 3h25 \\ 
 & \textbf{HN-I1.2} & 32 & 79 & 74 & - & 100\% & 1h45 \\ 
 & \textbf{HN-I1.4} & 23 & 79 & 10 & - & 100\% & 1h20 \\ 
 & \textbf{HN-I1.7} & 20 & 79 & 10 & - & 97\% & 0h20 \\ 
 & \textbf{HN-I2.0} & 20 & 79 & 11 & - & 92\% & 0h00 \\ 
 & \textbf{HN-I2.5} & 23 & 79 & 16 & - & 90\% & 0h00 \\ 
 & \textbf{HN-I3.0} & 29 & 79 & 24 & - & 84\% & 0h00 \\ 
 \bottomrule
\end{tabularx}
\caption{Experimental results for the \overt{FF} and \overt{CPG} models.}\label{tab:iub}
\end{table*}

\subsection{Validation through traffic simulation}
The optimization approaches presented in this work assume that the evacuees (or
vehicles) flow over the evacuation network in a continuous and aggregated
manner. In the real world however, evacuees are independent agents that move
along the edges and show different behaviors in response to the evacuation plan.
To assess the fitness and robustness of the results from the optimization, we
introduce an agent-based traffic simulation based on the MATSIM
\citep{MATSIM} simulation package.

In this simulation, each evacuee is modeled as an agent with an individual plan
composed by a start location (its evacuated area), a final destination (the
chosen safe node), a path in the evacuation graph, and a departure time. Each
individual plan can be either directly derived from the optimization results or
generated by introducing random perturbations. The MATSIM simulation engine uses
the set of plans to simulate the movement of evacuees in the evacuation graph.
It models each edge of the graph as a queue, which realistically simulates a
real-world transportation network, in particular by considering congestion.

The first simulation experiment we conducted aims at studying the feasibility of
the plan produced by the free flow model. Considering that FF does not produce a
plan for each evacuee, we considered two scenarios, namely \emph{Closest} and
\emph{Random Closest}. In the Closest scenario, each evacuee goes to the closest
accessible safe node at its departure time, and we ensure that the total volume
of evacuees leaving each area is the same as the one produced by the
optimization. In the Random Closest scenario, we allow for more variation in the
evacuees behaviors, and consider that 50\% will go to the closest, 40\% to one
of the five closest, and 10\% to a random safe node. In addition, we generate
random departure times that depend on the earliest departure time of the
neighboring areas and the latest departure time for the considered area. 

\begin{figure*}[htb]
 \includegraphics{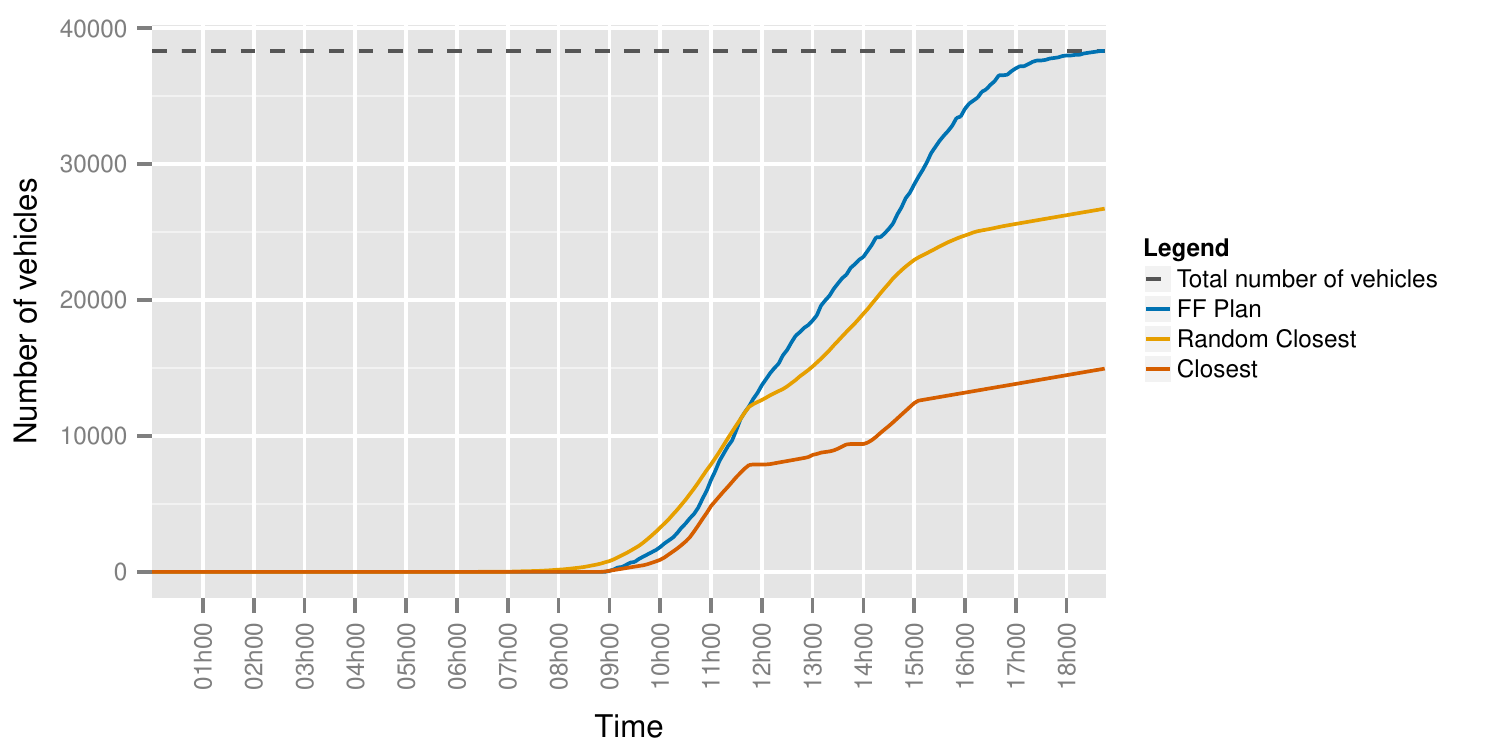}
 \caption{Comparison of evacuation profiles produced by the Free Flow (FF) model, Closest and Random Closest simulation scenarios on the HN instance.}\label{fig:evac_profile_ff}
\end{figure*}

Figure \ref{fig:evac_profile_ff} presents the evacuation profile produced by the
Free-Flow approach (FF Plan), and the Closest and Random Closest simulation
scenarios. The dashed line represent the total number of vehicles to be evacuated
in the considered area. This chart illustrates the
fact that FF schedule is too optimistic when simulated. More generally, our
experiments indicate that although FF predicts that all evacuees can reach a
safe zone in the HN and HN-Ix instances, the simulation indicates that only 39\%
would be evacuated by the end of the planning horizon in the Closest scenario,
and 66\% in the Random Closest scenario. Detailed results area available in
the Appendix.

The second experiment considers the evacuation plans produced by the path
generation approach. Table \ref{tab:sim-cg} presents the number of evacuees
reaching a safe node under different scenarios. The second column (Optimization
Plan) corresponds to the results of the optimization, and the third (Simulation
Plan) to the simulation of the plans produced by the optimization. The fourth
column (Rush) corresponds to a scenario in which all evacuees from a same area
leave at the first departure time produced by the optimization. The fifth column
(Random Schedule) is a scenario in which the departure time of the evacuees is
randomized depending on the earliest departure time of the neighboring areas and
the latest departure time for the considered area. The sixth column (Random
Plan) adds an additional level of randomization by considering that 50\% of the
evacuees will follow the plan, 40\% will go to one on the five closest safe
nodes, and 10\% will go to a random safe node. Finally, the seventh column
(Closest) represents a scenario in which evacuees depart as instructed but go to
the closest safe node. The results illustrate that the evacuation plan produced
by the CPG procedure is very close to the simulation and robust to evacuees
behaviors. Of particular interest is the fact that Simulation Plan, Rush, Random
Schedule, and Random Plan produce results within 3\% of what was predicted by
the Optimization Plan. In addition, our results indicate that CPG produces an
evacuation schedule that allow to evacuate a majority of evacuees even in the
Closest scenario.
\begin{table*}[h!]
\centering
\footnotesize
\begin{tabularx}{\textwidth}{lRRRRRR}
\toprule
\head{Instance} & \head{Opt. Plan} & \head{Sim. Plan} & \head{Rush} & \head{Rnd. Sched.} & \head{Rnd. Plan} & \head{Closest} \\ 
\midrule
\textbf{HN} & 100\% & 100\% & 100\% & 98\% & 97\% & 78\% \\ 
\textbf{HN-I1.1} & 100\% & 100\% & 100\% & 98\% & 95\% & 78\% \\ 
\textbf{HN-I1.2} & 100\% & 100\% & 100\% & 97\% & 97\% & 75\% \\ 
\textbf{HN-I1.4} & 100\% & 100\% & 100\% & 98\% & 95\% & 77\% \\ 
\textbf{HN-I1.7} & 97\% & 97\% & 98\% & 93\% & 96\% & 79\% \\ 
\textbf{HN-I2.0} & 92\% & 92\% & 93\% & 90\% & 93\% & 76\% \\ 
\textbf{HN-I2.5} & 90\% & 90\% & 91\% & 87\% & 90\% & 69\% \\ 
\textbf{HN-I3.0} & 87\% & 87\% & 87\% & 85\% & 89\% & 63\% \\ 
\bottomrule
\end{tabularx}
\caption{Percentage of evacuees reaching a safe node under different simulation scenarios.}\label{tab:sim-cg}
\end{table*}

\begin{figure*}[htb]
 \centering
 \includegraphics{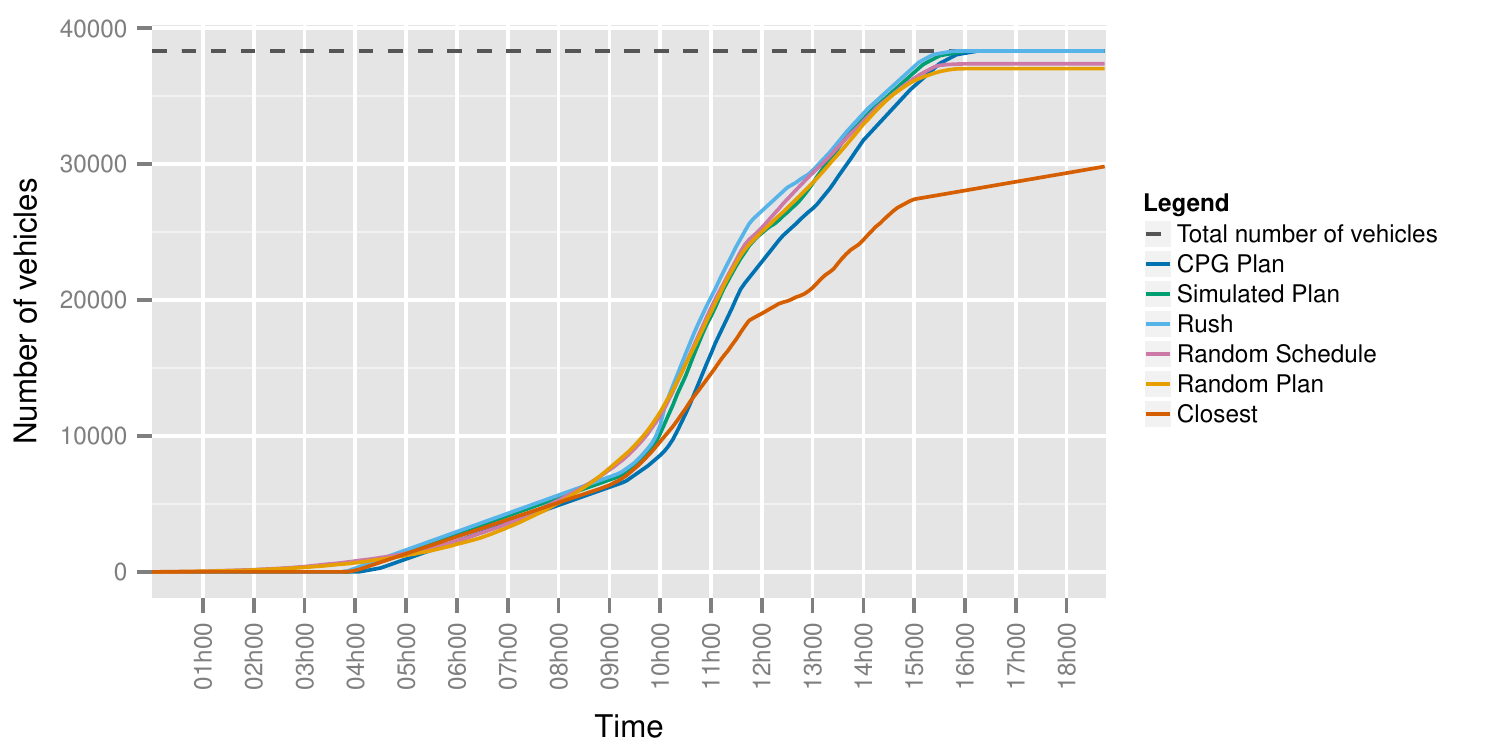}
 \caption{Comparison of evacuation profiles produced by the Path Generation (CPG) model, and different simulation scenarios on the HN instance.}\label{fig:evac_profile_cg}  
\end{figure*}

Figure \ref{fig:evac_profile_cg} illustrates the evacuation profiles produced by
the CPG approach and the different simulation scenarios for the HN instance.
These results illustrate the robustness of the plan produced by CPG: The curves
representing the number of evacuees reaching safety are very close independently
of the scenario considered. The only exception is Closest which generates more
congestion and for which only 78\% of the evacuees reach a safe node.

\section{Conclusions}\label{sec:conc}
This paper considered evacuation planning and scheduling, a critical
aspect of disaster management and national security applications. It
proposed a conflict-based path-generation approach whose
key idea is to generate evacuation paths for evacuated areas
iteratively and optimize the evacuation over these paths in a master
problem. Each new path is generated to remedy conflicts in the
evacuation and adds new columns and a new row in the master
problem.
The algorithm was applied to massive flood scenarios in the
Hawkesbury-Nepean floodplain (West Sydney, Australia) which require
evacuating about 70,000 persons. 
Computational results show that the proposed path-generation
approach is able to design evacuation plans for such large-scale scenarios in under 30
seconds, contrary to a traditional MIP approach which does not scale
to this problem size. Of particular interest is the fact that the
proposed approach reduces the number of variables from 4,500,000 in a
MIP formulation to 30,000 in the case study.

To the best of our knowledge, this is the first scalable evacuation
algorithm that conforms to evacuation methodologies and field
requirements. Our evacuation algorithm can be used in strategic,
tactical, and operational environments.

Our current work aims at
improving the path generation using constraint programming to find new
paths. Future work will also focus on microscopic modeling of the
transportation system, the inclusion of loading curves for
notification, and models of human behavior in evacuation settings.

\paragraph*{Acknowledgments}
NICTA is funded by the Australian Government as represented by the
Department of Broadband, Communications and the Digital Economy and
the Australian Research Council through the ICT Centre of1 Excellence
program.%

\appendix

\newpage

\pagestyle{special}
\markboth{APPENDIX}{}
\section*{Appendix: detailed simulation results for the free flow approach}\label{ap:res-ff}
Table \ref{tab:sim-ff} presents the percentage of evacuees reaching safety. The second column corresponds 
to the results of the optimization, and the third and fourth to the Closest and Random Closest simulation scenario. 
\begin{table*}[h!]
\centering
\footnotesize
\begin{tabularx}{.60\textwidth}{lRRR}
\toprule
\head{Instance} & \head{FF} & \head{Closest} & \head{Rnd. Closest} \\ 
\midrule
\textbf{HN} & 100\% & 39\% & 70\% \\ 
\textbf{HN-I1.1} & 100\% & 38\% & 68\% \\ 
\textbf{HN-I1.2} & 100\% & 38\% & 68\% \\ 
\textbf{HN-I1.4} & 100\% & 38\% & 67\% \\ 
\textbf{HN-I1.7} & 100\% & 37\% & 64\% \\ 
\textbf{HN-I2.0} & 100\% & 38\% & 63\% \\ 
\textbf{HN-I2.5} & 100\% & 40\% & 63\% \\ 
\textbf{HN-I3.0} & 100\% & 41\% & 63\% \\ 
\bottomrule
\end{tabularx}
\caption{Percentage of evacuees reaching a safe node when following the schedule produced by Free Flow (FF) under different simulation scenarios.}\label{tab:sim-ff}
\end{table*}

\newpage


\end{document}